\documentclass{article}
\usepackage{spconf,amsmath,graphicx,cite,booktabs,multirow,amssymb,enumitem,url}

\title{ADFA: Attention-augmented Differentiable top-k Feature Adaptation  for Unsupervised Medical Anomaly Detection}
\name{Yiming Huang$^{1,2}$, Guole Liu$^{1,2}$, Yaoru Luo$^{1,2}$, Ge Yang$^{1,2}$*\thanks{The work was supported in part by the National Natural Science Foundation of China (grants 31971289, 91954201, and 32101216) and the Strategic Priority Research Program of the Chinese Academy of Sciences (grant XDB37040402).}}
\address{$^1$State Key Laboratory of Multimodal Artificial Intelligence Systems, \\Institute of Automation, Chinese Academy of Sciences, Beijing, 100190, China\\
$^2$School of Artificial Intelligence, University of Chinese Academy of Sciences, Beijing, 100049, China}

\begin{document}
%
\maketitle
\begin{abstract}
The scarcity of annotated data, particularly for rare diseases, limits the variability of training data and the range of detectable lesions, presenting a significant challenge for supervised anomaly detection in medical imaging. To solve this problem, we propose a novel unsupervised method for medical image anomaly detection: Attention-Augmented Differentiable top-k Feature Adaptation (ADFA). The method utilizes Wide-ResNet50-2 (WR50) network pre-trained on ImageNet to extract initial feature representations. To reduce the channel dimensionality while preserving relevant channel information, we employ an attention-augmented patch descriptor on the extracted features. We then apply differentiable top-k feature adaptation to train the patch descriptor, mapping the extracted feature representations to a new vector space, enabling effective detection of anomalies. Experiments show that ADFA outperforms state-of-the-art (SOTA) methods on multiple challenging medical image datasets, confirming its effectiveness in medical anomaly detection.
\end{abstract}

\section{Introduction}
\label{sec:intro}
Anomaly detection (AD) in medical images is an important task, as it can provide earlier diagnosis, better patient outcomes, and a better understanding of underlying pathology. For example, mammograms and other imaging tests are commonly used for the early detection of breast cancer~\cite{KOOI2017303}. In previous studies, the application of supervised learning techniques for the identification of lesions in clinical images has generated impressive results, achieving accuracy levels comparable to those of experienced clinical experts~\cite{KOOI2017303,liu2022data}.
Supervised AD depends heavily on the availability of accurate and well-labeled training data. However, obtaining such data can be a difficult, time-consuming, and expensive task, especially for rare diseases like hereditary spastic paraplegia, which has an incidence rate as low as 1.27 per 100,000. Furthermore, the visual variability in the annotated training data restricts the range of lesions that can be detected by supervised AD~\cite{fanogan}. As an alternative approach, unsupervised AD in medical images has gained considerable attention~\cite{minireview}. It requires only normal images for training and is capable of identifying abnormal patterns or structures in medical images.

\begin{figure}[t]
    \centering
    \includegraphics[width=0.48\textwidth]{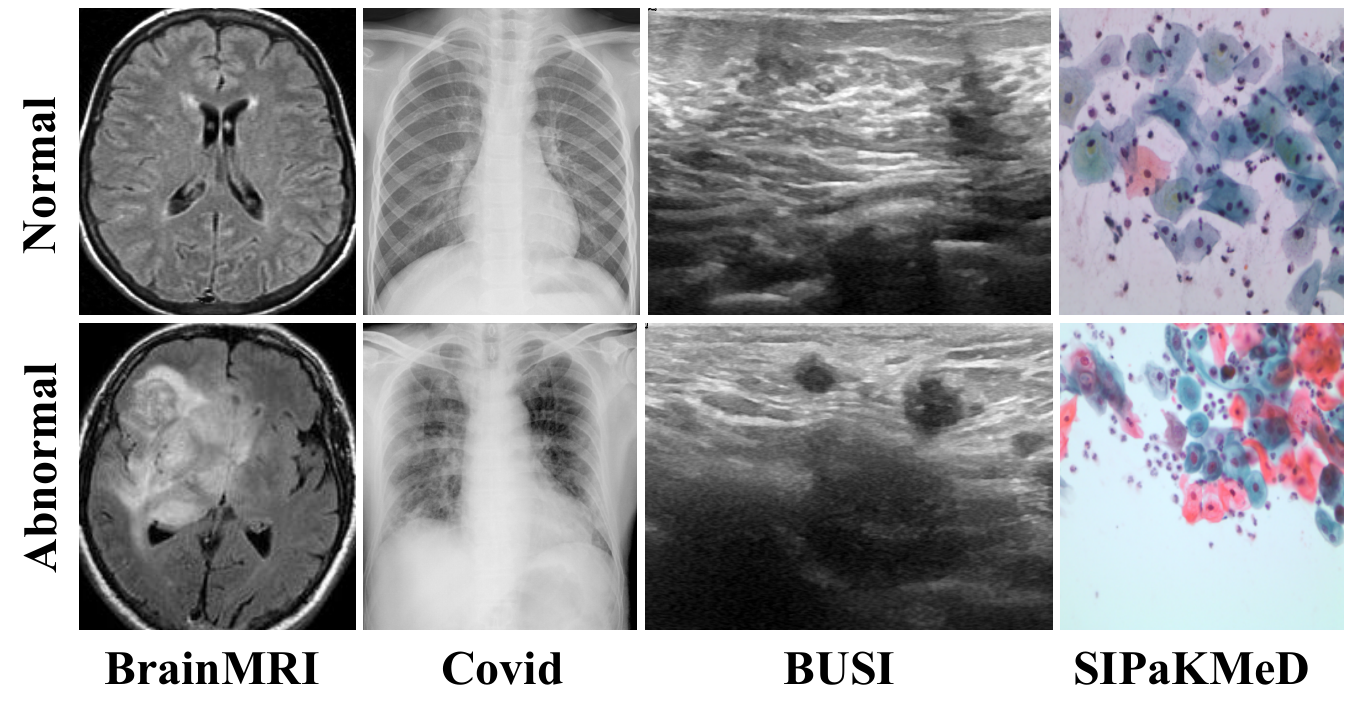}
	\caption{\textbf{Examples of normal and abnormal images from four datasets~\cite{KDAD,covid,BUSI,SIPaKMeD}.} }
	\label{images}
\end{figure} 

\begin{figure*}[t]
    \centering
    \includegraphics[width=\textwidth]{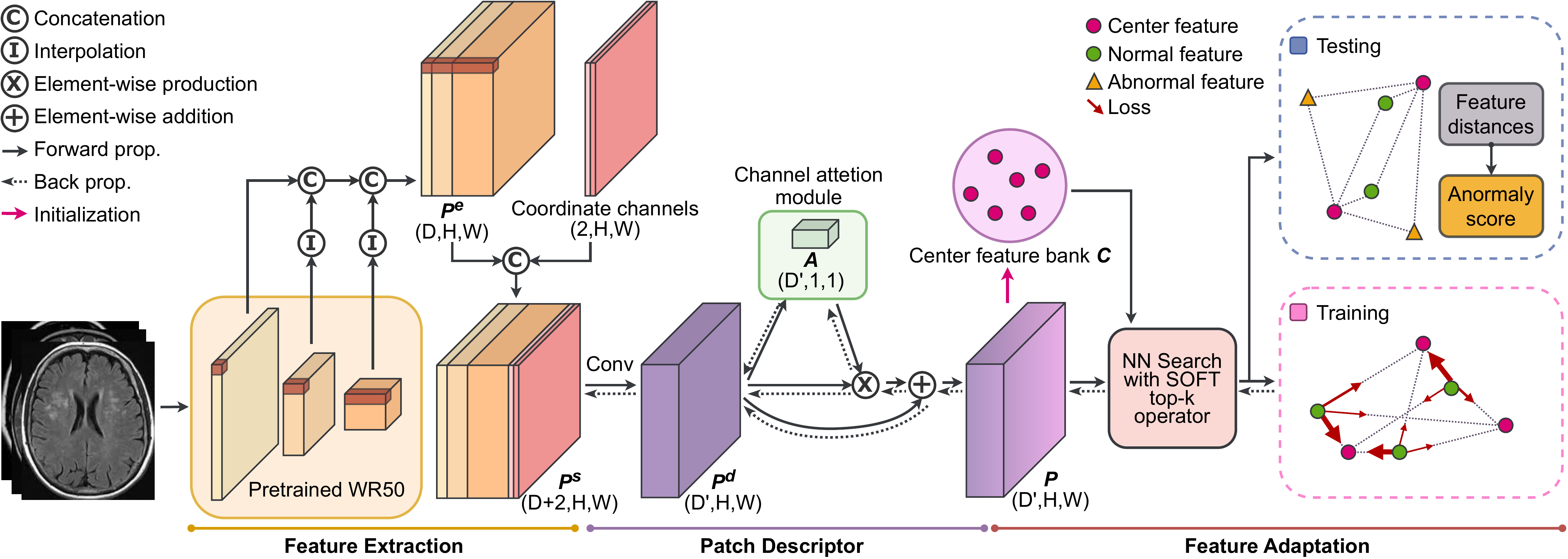}
	\caption{\textbf{An overview of the ADFA method.} It consists of feature extraction, patch descriptor, and feature adaptation.}
	\label{fig1}
\end{figure*}  
Numerous unsupervised methods for AD have been proposed, which can be divided into three categories.  \textbf{1) Reconstruction based methods}, which learn the latent space representation of normal data, and identify abnormal data by their reconstruction errors. Advances in this area include the use of Generative Adversarial Networks (GANs)~\cite{anogan,fanogan} and Variational Autoencoders (VAEs)~\cite{gmvae}. 
\textbf{2) Knowledge distillation (KD) based methods}~\cite{student,kduninformed,KDAD,rdad}, which utilize a student-teacher network to train the student network only on normal data and then identify abnormal data through feature inconsistencies. 
\textbf{3) Embedding similarity based methods}, which employ deep neural networks to generate feature vectors for the entire image or patches of an image. Anomaly scores are calculated based on the distance between the feature vectors of a test image and the normal reference, which can be the center of a hypersphere containing embeddings ~\cite{deepone,patchsvdd,cfa}, parameters of Gaussian distributions~\cite{mad,padim} or the entire set of normal embedding vectors~\cite{spade,bergman2020deep}.

Overall, most unsupervised medical AD studies have utilized reconstruction as a central approach~\cite{review}. These methods are intuitive and interpretable. However, generative models sometimes do accurately reconstruct abnormal regions, leading to difficulties in detecting anomalies~\cite{padim}. KD based and embeddings similarity based methods perform well on industrial datasets such as MVTecAD~\cite{mvtec}. Nevertheless, their application to medical data is limited~\cite{review}.

In this work, a novel unsupervised method for medical image AD is proposed. The method starts with using a Wide-ResNet50-2 (WR50)~\cite{wrn} network pre-trained on ImageNet dataset as a feature extractor and then concatenates feature maps from different layers to capture both fine-grained and global context information. The extracted features are then concatenated with the 2D coordinates of each pixel to obtain embedding vectors. As embedding vectors carry redundant information for AD tasks, a patch descriptor composed of 2D convolution with a channel attention module is then applied to the features to reduce channel dimensionality while retaining important channel information. The final step applies a differentiable top-k operator to train the patch descriptor, which maps the extracted features to a new space for improved anomaly detection accuracy. This allows our approach to capture complex relationships between image features and gather feature vectors of normal images more closely.

The major contributions of this paper are as follows:
\begin{itemize}
\setlength{\itemsep}{0pt}
\setlength{\parsep}{0pt}
\setlength{\parskip}{0pt}
    \item [1)]
    We propose a novel unsupervised AD model\footnote{Code is available at \url{https://github.com/cbmi-group/ADFA.git}} based on feature adaptation by utilizing a pre-trained feature extractor to handle the challenges posed by the scarcity of annotated medical imaging data. It outperforms SOTA methods in average performance on four medical image datasets.
    \item [2)]
    Our model uses an attention-augmented module and a differentiable top-k operator to improve the separability of normal and abnormal image features, enhancing the ability to detect anomalies in complex medical imaging data with high heterogeneity.
\end{itemize}

\section{METHOD}
\label{sec:format}
The proposed method, as shown in Fig.~\ref{fig1}, comprises three sub-modules: feature extraction, patch descriptor, and feature adaptation, which will be detailed in the following sections.
\subsection{Feature extraction}
\label{ssec:subhead}
Previous research by Bergman et al.~\cite{bergman2020deep} has shown that pre-trained CNNs are able to output relevant features for AD. In this work, we use a WR50 pre-trained on ImageNet as a feature extractor. In order to capture both fine-grained and global context information, feature vectors from different layers are interpolated to the same resolution and then concatenated. Given an input image $x$, features \textbf{$P^{e}$}$\in \mathbb{R} ^{D\times H\times W}$ can be extracted from it using the feature extractor, where $D$ indicates the sum of channels of the sampled feature maps, and $H\times W$ means the resolution of the largest feature map.

Incorporating the spatial context of an image into convolutional filters has been demonstrated to enhance the robustness of learned features~\cite{coordconv}. Given the significance of spatial context in medical imaging, we integrate the 2D coordinates of each pixel as supplementary channels to the feature vectors, thereby augmenting the model's spatial awareness. Specifically, we add row and column coordinate channels and scale their values linearly to the range [-1, 1]. We refer to the resulting feature vectors as spatially-aware features $P^s\in\mathbb{R} ^{(D+2)\times H\times W}$.

\subsection{Attention-augmented patch descriptor}
\label{ssec:subhead}
Features generated from pre-trained CNNs carry redundant information that is not useful for AD tasks~\cite{padim}. To address this, we utilize a patch descriptor with a channel attention module that helps to reduce the channel dimensionality of the features while preserving the essential channel information.

\begin{figure}[t]
    \centering
    \includegraphics[width=0.48\textwidth]{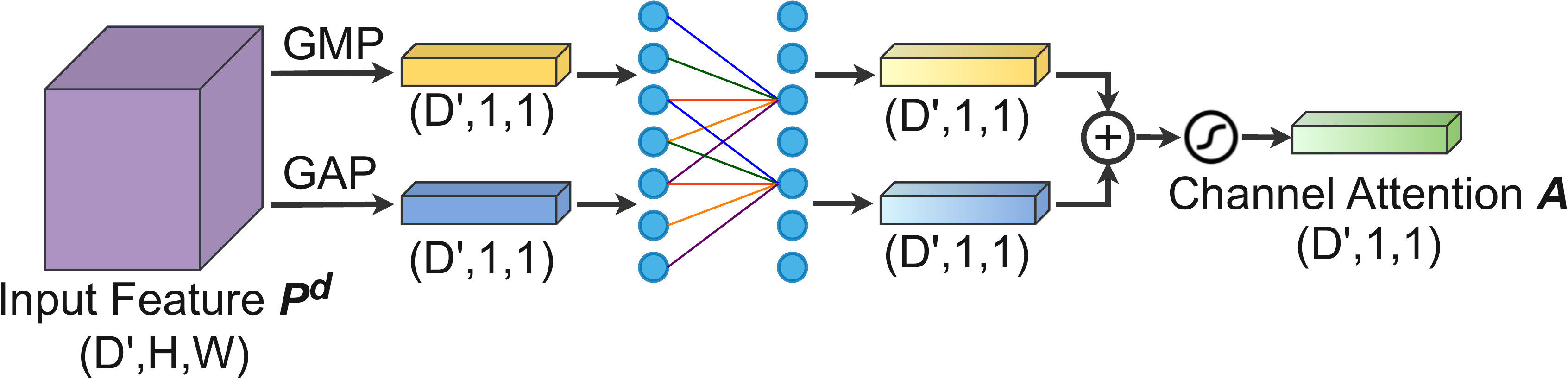}
	\caption{\textbf{Diagram of channel attention module.} The module utilizes both GMP and GAP outputs with a shared 1D convolution layer.}
	\label{fig2}
\end{figure}

After obtaining the spatially-aware features $P^s$, we apply a $1\times 1$ 2D convolutional layer to obtain low-dimensional features $P^d\in\mathbb{R} ^{D'\times H\times W}$. We further process $P^d$ through a channel attention module, which is depicted in Fig.~\ref{fig2}. 
The channel attention module aggregates spatial information of the input feature map using both global max pooling (GMP) and global average pooling (GAP) operations, which improves the representation power of the module when used together rather than separately~\cite{woo2018cbam}. The output feature vectors of the pooling operations are forwarded to a shared 1D convolution $C1D_k$, where kernel size $k$ is adaptively determined by channel dimensions $D'$~\cite{wang2020eca}. These vectors are then combined element-wise using summation, and the resulting vector is passed through a sigmoid activation function ($\sigma$) to generate the final channel attention map $A\in\mathbb{R} ^{D'\times 1\times 1}$:
\begin{equation}
A(P^d) = \sigma(C1D_k(GMP(P^d))+C1D_k(GAP(P^d))).
\end{equation}
Finally, refined features $P\in\mathbb{R}^{D'\times H\times W}$ can be obtained: 
\begin{equation}
P = P^d + \varepsilon( A(P^d)) \otimes P^d), 
\end{equation}
where $\varepsilon$ is a hyper-parameter, and $\otimes$ denotes element-wise multiplication. During multiplication, the  channel attention values are broadcasted along the spatial dimension. By flattening $P$, we can obtain patch feature vectors $p_{t\in{1,...,HW}}\in\mathbb{R}^{D'}$, each of which corresponds to a patch of the input image $x$ with a specific receptive field.

This module significantly reduces the computational complexity and memory requirements of the model, while still capturing significant contextual information from the spatially-aware features, leading to impressive performance.

\subsection{Differentiable top-k feature adaptation}
\label{ssec:subhead}
\label{sec2.3}
This section describes the process of training the patch descriptor, that maps patch features to a new vector space. Patches have high variation within one image and some may contain the object while others may be background, thus using a single center to represent all patches is not suitable~\cite{patchsvdd}. Therefore we use a center feature bank $\mathcal{C}$ to store center feature vectors. For all patch feature vectors $p_t^i$ obtained from normal training samples $x^i \in \mathcal{X}^N$ , $\mathcal{C}$ is defined as
\begin{equation}
\mathcal{C}=\left \{c_t \mid c_t = \frac{1}{N} \sum_{i=1}^{N} p^i_t , t\in{1,...,HW} \right \},
\end{equation}
where center feature vectors $c_{t\in{1,...,HW}}\in\mathbb{R}^{D'}$ are initialized by computing the average of patch feature vectors.

During the training phase, we utilize nearest neighbor (NN) search with SOFT top-k operator~\cite{soft}, which is differentiable and can be optimized using gradient-based optimization techniques, to find the $K$-nearest centers for each patch feature vector $p_t^i$. First, the vector $\mathcal{D}(p_t^i,\mathcal{C})$ consisting of the distances from $p_t^i$ to the set of representative centroids $\mathcal{C}$ is computed using the Euclidean distance:
\begin{equation}
\mathcal{D}(p_t^i, \mathcal{C})=\begin{bmatrix}\left\|p_t^i-c_1\right\|_{2},\ldots,\left\|p_t^i-c_{HW}\right\|_{2}\end{bmatrix}.
\end{equation}
The SOFT top-k operator is then applied to this vector to obtain the $K$-smallest distances. The resulting vector $\mathcal{Z}_t$ contains values between 0 and 1 that represent the likelihood that the distance is among the $K$-smallest. The loss function of ADFA is given by the equation:
\begin{equation}
\mathcal{L}_{ADFA} =\frac{1}{HW} \sum_{t=1}^{H W} \mathcal{Z}_t^\top  \mathcal{D}(p_t^i, \mathcal{C}).
\end{equation}
ADFA supervises patch descriptor by minimizing $\mathcal{L}_{ADFA}$ so that $p_t^i$ is embedded close to the $K$-nearest centers. Specifically, the objective is to adapt the pre-trained feature to the smallest hyperspheres that encompass the normal training data in the kernel space, with the expectation that any anomalies will be located outside of these learned hyperspheres.

\begin{table*}[t]
\centering
\begin{tabular}{c|cccccc|c}
\hline
Datasets  & PaDiM\cite{padim} & SPADE\cite{spade} & Patch SVDD\cite{patchsvdd} & STPM\cite{student}  & RDAD\cite{rdad}           & f-AnoGan\cite{fanogan} & ADFA(\textbf{Ours})   \\ \hline
BrainMRI & 0.843 & 0.682 & 0.647 & 0.821 & \underline{0.856} & 0.733    & \textbf{0.857} \\
Covid    & 0.695 & 0.680  & 0.918 & 0.757 & 0.908 & \textbf{0.982}    & \underline{0.973} \\
BUSI     & 0.948 & \underline{0.945} & 0.751 & 0.796 & 0.862          & 0.785    & \textbf{0.966} \\
SIPaKMeD & 0.754 & 0.896 & \underline{0.935} & 0.792 & 0.784          & 0.828    & \textbf{0.972} \\ \hline
\end{tabular}
\caption{Performance of our model and related work on four datasets (AUROC\%). Bold: best result. Underline: second best.}
\label{tab:main_result}
\end{table*}

\begin{table}[t]
\centering
\resizebox{0.7\linewidth}{!}{
\begin{tabular}{c|c|cc}
\hline
\multirow{2}{*}{Datasets} & Training & \multicolumn{2}{c}{Testing} \\ \cline{2-4} 
                          & Normal   & Normal      & Abnormal     \\ \hline
BrainMRI                  & 58       & 40          & 155          \\
Covid                     & 74       & 20          & 20           \\
BUSI                      & 96       & 32          & 496          \\
SIPaKMeD                  & 180      & 54          & 461          \\ \hline
\end{tabular}}
\caption{Key statistics of image datasets.}
\label{tab:datasets}
\end{table}

\subsection{Anomaly scoring}
\label{ssec:subhead}
In the testing phase, we need to get the anomaly score of the input image. As stated in Section~\ref{sec2.3}, we use SOFT top-k operator to calculate the $K$-smallest distance between $p_t$ and the center features $\mathcal{C}$. So, we can define the anomaly score of a sample $x$ naively as shown below:
\begin{equation}
\mathcal{S}(x) =\max \{ \mathcal{Z}_t^\top  \mathcal{D}(p_t, \mathcal{C})\}.
\end{equation}
We designate the maximum value in sample $x$ as the anomaly score. This approach was chosen because the size of anomalous regions in medical images can vary widely, and using the average value as the anomaly score would be unfair.

\section{EXPERIMENTS}
\label{sec:pagestyle}

\subsection{Datasets and metrics}
\label{ssec:subhead}
We evaluate our model on four medical datasets: \textbf{BrainMRI}~\cite{KDAD}, a dataset including normal brain MRI images and those with brain tumors; \textbf{Covid}~\cite{covid}, a dataset of chest X-ray images of both healthy individuals and COVID-19 patients; \textbf{BUSI}~\cite{BUSI}, a dataset composed of chest ultrasound images of women between the ages of 25 and 75, both healthy and with breast cancer.; \textbf{SIPaKMeD}~\cite{SIPaKMeD}, a dataset of cervical cells in Pap smear images, including 4 categories. The number and the division of datasets are shown in Table~\ref{tab:datasets}.

Performance of the AD method is evaluated using the Area Under the Receiver Operating Characteristic curve (AUROC), with a value of 1 indicating optimal performance and values close to 0.5 indicating random classification.

\begin{table}
\centering
\resizebox{\linewidth}{!}{
\begin{tabular}{c|cccc|c|c}
\hline
\multirow{2}{*}{Datasets} & \multicolumn{4}{c|}{$\varepsilon$}                                  & \multirow{2}{*}{\begin{tabular}[c]{@{}c@{}}PyTorch\\ top-k\end{tabular}} & \multirow{2}{*}{\begin{tabular}[c]{@{}c@{}}randomly \\ initialized\end{tabular}} \\ \cline{2-5}
                          & 0.00           & 0.05           & 0.10           & 0.20        &                                                                          &                                                                                  \\ \hline
BrainMRI                  & \textbf{0.858} & \textbf{0.858} & \underline {0.857}    & 0.855       & 0.844                                                                    & 0.577                                                                                \\
Covid                     & 0.963          & 0.967          & \textbf{0.973} & \underline{0.970} & 0.917                                                                    & 0.470                                                                                 \\
BUSI                      & 0.958          & 0.962          & \textbf{0.966} & \underline{0.965} & 0.952                                                                    & 0.591                                                                                 \\
SIPaKMeD                  & 0.964          & \underline{0.971}    & \textbf{0.972} & \textbf{0.972}       & 0.958                                                                    & 0.512                                                                                 \\ \hline
\end{tabular}}
\caption{Ablation experiment results (AUROC\%).}
\label{tab:ablation}
\end{table}

\subsection{Experimental setups}
\label{ssec:subhead}
We resize all images to $256\times256$ and center-crop them to $224\times224$. We use the WR50 model pre-trained on the ImageNet dataset to extract feature maps from the first three layers of the network. The spatial resolution of each feature map is 1/4, 1/8, and 1/16 of the input image, respectively, and their corresponding channel dimensions are 256, 512, and 1024. Spatially-aware features with coordinate channels $P^s\in \mathbb{R}^{1794\times 56\times 56}$ and low-dimensional features $P^d\in\mathbb{R}^{448\times 56\times 56}$.  To optimize the patch descriptor parameters, we use the AdamW optimizer with AMSGrad and set the learning rate to 1e-3 without any learning rate scheduler. The weight decay is set to 5e-4. We set the hyper-parameters of ADFA, namely $K$ and $\varepsilon$, to 3 and 0.1, respectively. We perform all our experiments on a single NVIDIA GeForce RTX 3090 GPU.

\subsection{Results and discussions}
\label{ssec:subhead}
The experiment results in Table~\ref{tab:main_result} compare the performance of the proposed ADFA method with several SOTA AD methods on image datasets. It can be seen that ADFA outperforms the other methods, or is comparable to the best-performing method. It is also worth noting that the results on specific medical image datasets also show that ADFA performs well on different types of data. ADFA achieves the highest scores on the BUSI and SIPaKMeD datasets, having an AUROC improvement of 2.1\% and 3.7\%, respectively. This demonstrates the robustness and versatility of the proposed method in detecting anomalies in medical images.

\subsection{Ablation study}
    \label{ssec:subhead}
Table~\ref{tab:ablation} presents the results of an ablation study to evaluate the contribution of the different components of our method. The study focuses on validating the effectiveness of the channel attention module, which we do by testing the model with different values of $\varepsilon$. The absence of the attention module, represented by $\varepsilon=0$, results in poor performance. On the other hand, we find that setting $\varepsilon=0.1$ produces the best overall performance.

Additionally, we examine the impact of using the SOFT top-k operator versus the non-differentiable top-k operator from PyTorch, as well as the effect of using a randomly initialized WR50 instead of a pre-trained one. The results indicate that both of these alternatives perform worse than our proposed model, highlighting the importance of the attention module and differentiable top-k operator for achieving high performance. Thus, we can conclude that our method is effective in detecting anomalies in medical images.

\section{CONCLUSION}
\label{sec:typestyle}

We propose a novel method called ADFA for AD in medical imaging. It leverages pre-trained WR50 as a feature extractor and uses differentiable top-k feature adaptation to learn an attention-augmented patch descriptor, allowing it to effectively identify anomalies in the images. It is tested on four medical image datasets and outperforms SOTA methods in average performance. ADFA shows great potential for practical applications in the field of medical image analysis. Future work will focus on implementing anomaly localization and exploring the potential of ADFA for other applications.


%
\newpage

\ninept
\bibliographystyle{IEEEbib}
\bibliography{strings,refs}

\end{document}